\documentclass[10pt,twocolumn,letterpaper]{article}

\usepackage{cvpr}
\usepackage{times}
\usepackage{epsfig}
\usepackage{graphicx}
\usepackage{amsmath,bm}
\usepackage{amssymb}
\usepackage{hyperref}      
\usepackage{url}           
\usepackage{booktabs}      
\usepackage{amsfonts}      
\usepackage{nicefrac}      
\usepackage{microtype}     
\hypersetup{colorlinks=true, linkcolor=red}
\usepackage{pgfplots}
\usepackage{pgfplotstable}
\usepackage{filecontents,pgfplots}
\usepackage{tikz}
\usetikzlibrary{arrows,calc,shapes,snakes,positioning,matrix,arrows,decorations.pathmorphing,decorations.text}
\usepackage{mathrsfs}
\usepackage{soul}
\usepackage{multirow}
\usepackage{comment}
\usepackage{rotating}
\usepackage{mwe}
\PassOptionsToPackage{bookmarks=false}{hyperref}

\newcommand{\DD}{\mathcal{D}}
\newcommand{\II}{\mathcal{I}}
\newcommand{\CC}{\mathcal{C}}
\usepackage[skip=1ex]{caption}
\usepackage{array, booktabs, makecell, multirow}
\setcellgapes{5pt}
\cvprfinalcopy

\begin{document}

\title{G2D: Generate to Detect  Anomaly}

\author{Masoud Pourreza$^1$, Bahram Mohammadi$^2$, Mostafa Khaki$^3$, \\ Samir Bouindour$^4$, Hichem Snoussi$^4$, Mohammad Sabokrou$^5$\\
$^1$Towzin LTD AI Group, $^2$Sharif University of Technology, $^3$Linkpey AI Group \\ $^4$University of Technology of Troyes, $^5$Institute For Research In Fundamental Sciences (IPM)}

\maketitle

\begin{abstract}
	In this paper, we propose a novel method for irregularity detection. Previous researches solve this problem as a One-Class Classification (OCC) task where they train a reference model on all of the available samples. Then, they consider a test sample as an anomaly if it has a diversion from the reference model. Generative Adversarial Networks (GANs) have achieved the most promising results for OCC while implementing and training such networks, especially for the OCC task, is a cumbersome and computationally expensive procedure. To cope with the mentioned challenges, we present a simple but effective method to solve the irregularity detection as a binary classification task in order to make the implementation easier along with improving the detection performance. We learn two deep neural networks (generator and discriminator) in a GAN-style setting on merely the normal samples. During training, the generator gradually becomes an expert to generate samples which are similar to the normal ones. In the training phase, when the generator fails to produce normal data (in the early stages of learning and also prior to the complete convergence), it can be considered as an irregularity generator. In this way, we simultaneously generate the irregular samples. Afterward, we train a binary classifier on the generated anomalous samples along with the normal instances in order to be capable of detecting irregularities. The proposed framework applies to different related applications of outlier and anomaly detection in images and videos, respectively. The results confirm that our proposed method is superior to the baseline and state-of-the-art solutions.
\end{abstract}


\section{Introduction}
\label{sec:introduction}
One-Class Classification (OCC) is the task of detecting samples which are unseen or out-of-target distribution. In other words, an OCC method looks for anomalies, \ie, unexpected behaviors or events which do not (or rarely) occur in the training data \cite{tang2020integrating}. Such methods should be trained just in the presence of normal samples (in distribution) and do not mind the concept of outlires. Furthermore, outliers are conceptually diverse and very complex. These difficulties alongside the high correlation of OCC to a wide range of computer vision and machine learning applications pose an ongoing challenge.

Inspired by the successes of deep learning solutions, especially Generative Adversarial Networks (GANs), some methods like ALOCC \cite{sabokrou2018adversarially,sabokrou2020deep} have been presented for outlier detection. In a nutshell, such methods adversarially learn two deep neural networks, one for generating fake data and another for discriminating normal samples from outliers. Although they have achieved promising results, as same as all other GAN-based methods, learning to find the optimum parameters is a bothersome procedure. Also, for the OCC task, finding the appropriate time for stopping the learning process to achieve the best performance without any validation samples of outlier class is very challenging and needs trial and error \cite{sabokrou2018adversarially,sabokrou2018avid}. In beside of this, their performance is highly dependent on hyper-parameters such as number of layers, kernels and the learning rate. 

\begin{figure}[t]
	\centering
	\includegraphics[width=1\columnwidth]{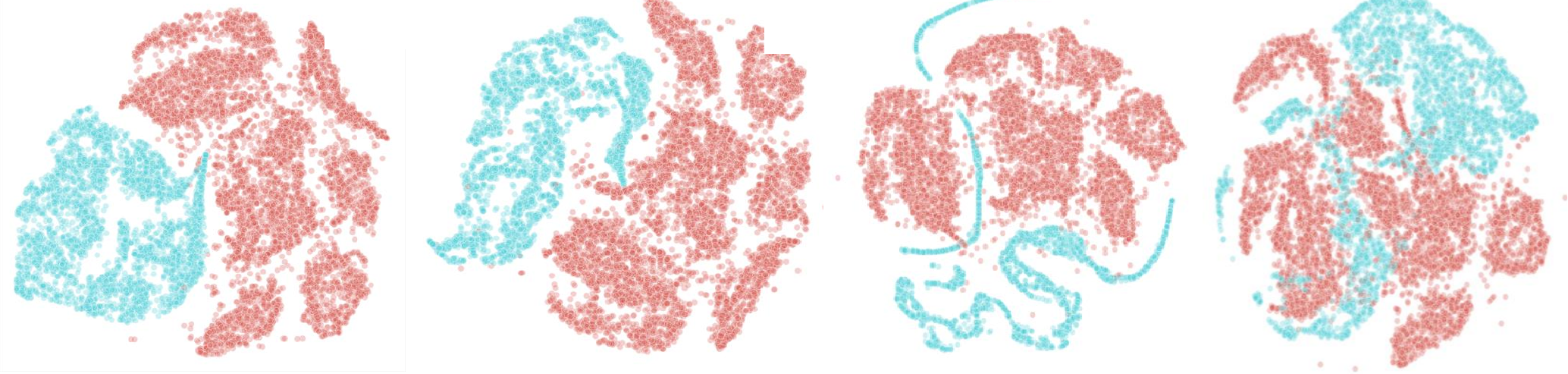}
	\caption{Generated instances (blue) by a GAN trained on normal (target class) samples (red) during training in different epochs. We use t-SNE to represent samples in two dimensions. As can be seen in earlier epochs, the generator of GAN produces samples which are completely different from normal ones. But as the learning process continues by the increase of epochs, both types of samples are getting close to each other gradually.}
	\label{fig:tsne}
\end{figure}{}

To address the mentioned challenges, we propose a very simple yet efficient method for detecting outlier samples. Similar to the prior approaches, we take advantage of the adversarial learning, but use it in a completely different manner. Previously proposed methods such as AVID \cite{sabokrou2018avid} and ALOCC \cite{sabokrou2018adversarially}, have focused to adversarially learn the distribution of normal data and consider all of the instances with a major diversion from the reference model as outlier. In contrary to such solutions, we suggest a simple and straightforward way for generating irregular samples which do not follow the distribution of normal data. Generated irregularities alongside of available normal instances, simply can be used for training a binary classifiers.  

GAN is a well-known tool for generating samples with the same distribution of training ones, \ie, normal class data. Having analyzed the generator network of a GAN during the training procedure, we have interestingly found that its role is changed as long as the learning process is not finished, especially when the GAN is trained only on the target class data for the OCC task. Regarding this analysis, we decide to benefit from the training process of the GAN for generating outliers. In Fig.~\ref{fig:tsne}, the generated samples (blue) using GAN (\ie, generator network trained on target samples colored red) in different epochs of training are shown. As can be seen, they successfully generate the samples near the target class, but as outliers. By accessing to the target class data and outliers surrounding them (boundary samples), a binary classifier can be simply and efficiently used to make discrimination between target and irregular class. Availability of data from both target and outlier class in the training phase, empowers our algorithm to be almost insensitive to the change of irregularities in number, in the stage of test. In fact, our method does experience a considerable drop in performance in case that irregular samples increase.

The main contributions of this paper are: (1) we propose an effective solution utilizing GAN for generating unseen (abnormal) samples. To the best of our knowledge, our proposed model for generating irregular data is the first method in this area of research, (2) most of the previous approaches for the OCC task which is based on the adversarial learning, are very difficult for training and suffer from GANs weaknesses such as instability. In our case, to avoid these challenges, we exploit GANs to generate abnormal samples (outliers) in a completely different manner. Having generated outliers, we simplify the OCC problem and convert it into a binary classification task, (3) learning the concept of normal and outlier class data helps our algorithm to operate more robust compared to the other approaches, and (4) our method is merely trained on samples belonging to the target class. Furthermore, we have achieved the state-of-the-art performance in different applications, such as outlier detection in images and anomalous event detection in videos.

\section{Related Works}
\label{sec:related_work}
OCC is an umbrella term for a group of tasks such as rare event detection, outlier detection/removal and anomaly detection. All of these tasks are looking for a concept that does not (or scarcely) occur in the training data. Consequently, many of real-world problems are highly correlated to the OCC task. Traditional OCC methods learn  the concept of the target class data as a normal reference model. Afterward, the samples with a high deviation from this model are detected as novelty.   

Self-representation learning \cite{xia2015learning,you2017provable,cong2011sparse,sabokrou2016video} and statistical modeling \cite{markou2003novelty} are two widely used approaches for solving the OCC task. Representing of video or image data using a set of features is a fundamental pre-process. The most used features for representing the images or videos for the OCC task are: (1) low level features \cite{bertini2012multi}, (2) high level features (\eg, trajectories \cite{morris2011trajectory}), and (3) deeply learned features \cite{xu2015learning,sabokrou2015real,sabokrou2017deep}. In the following, we survey the effective and successful methods have been proposed for OCC problem, by concentrating on those inspired by the generative adversarial learning.

\noindent\textbf{Self-Representation}
As has been investigated by the previously proposed methods, self-representation is a worthwhile approach for novelty detection task. For instance, Cong \etal~\cite{cong2011sparse} and Sabokrou \etal~\cite{sabokrou2016video} exploit the self-representation to detect the irregular events in video and also novelty by taking the advantages of sparse representation learning. In fact, in these works the sparsity is utilized as a criterion for distinguishing between inlier and outlier samples. Furthermore, in the test phase of the some methods, like what has been suggested by Xu \etal~\cite{xu2015learning} and Cong \etal~\cite{cong2011sparse}, samples are reconstructed by an encoder-decoder neural network trained on the target class data. In this case, the decision about the type of a sample is made based on the reconstruction error. In other words, if the reconstruction error for a sample becomes less than a threshold, it is considered as inlier. Otherwise, \ie, the reconstruction error exceeds the threshold, the sample is detected as outlier. Liu \etal~\cite{liu2010robust} used a low-rank representation rather than  sparse representation and penalized by the sum of self-representation errors. It is worth mentioning that, this penalization led to more robustness against outliers (analogous with the work of Adeli \etal~\cite{adeli2015robust}). Similarly, Sabokrou \etal~\cite{sabokrou2019self,sabokrou2016video} and Xu \etal~ \cite{xu2015learning} proposed algorithms in which the encoder-decoder neural networks are trained based reconstruction error loss function exploited for measuring outlier removal and video anomaly detection.  In \cite{nguyen2019anomaly} a method based on modeling the correspondence between common object appearances and their associated temporal features is presented. 

\begin{figure*}[t]
	\includegraphics[width=1\linewidth]{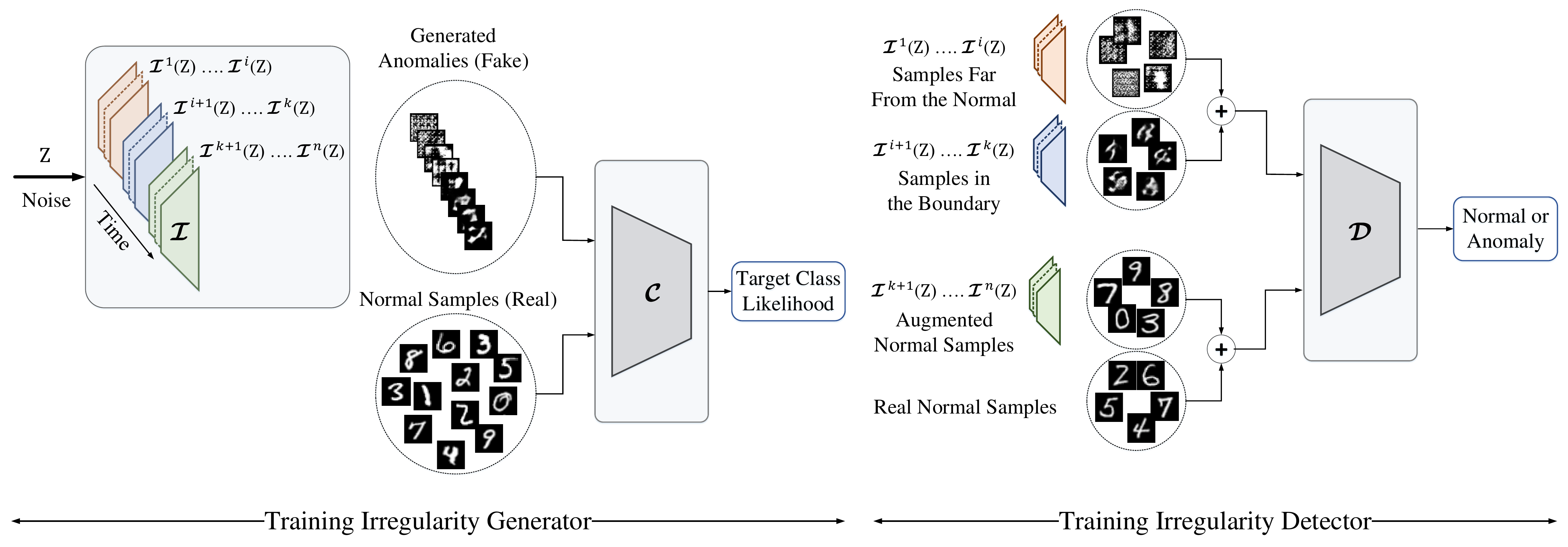}
	\caption{The outline of our proposed method. (\textit{Left}:) $\II$rregularity generator, (\textit{Right}:) Irregularity $\DD$etector. $\II+\CC$ networks are jointly and adversarially trained on normal class data. In training duration, several models ($\II$ with different weights) are considered as the irregularity detector. The $\DD$ network as a binary classifier is trained on all of the available normal samples and generated irregular samples by $\II^{1:k}$. After the training process, $\DD$ acts as an irregularity detector. }
	\label{fig:wa}
\end{figure*}

\noindent\textbf{Statistical Modeling}
Several works attempt to understand the positive samples by analyzing their statistical characteristics. As a simple way, they represented the samples from the target class to a reduced dimension feature space, a probability distribution with the maximum likelihood  is fit on such represented samples, after that, samples that do not comply the fitted distribution are detected as outliers or novelty (\eg, \cite{eskin2000anomaly, yamanishi2004line, markou2003novelty}). \cite{rahmani2016coherence} an efficient method named Coherence Pursuit (CoP) for Robust Principal Component Analysis (RPCA) is proposed by Rahmani and Atia. This method assumed that the correlation between the  inlier samples is very high, and can be spanned in low dimensional sub-spaces, and thus they have a strong mutual coherence with a large number of data points. As a result, the outliers either do not accord with the low dimensional subspace or form small clusters. Also, a method proposed in \cite{xu2010robust}, OutlierPursuit, used convex optimization techniques to solve the PCA problem with robustness to corrupted entries, which led to the development of many recent methods for PCA with robustness to outliers. Lerman \etal~\cite{lerman2015robust} introduced REAPER, a convex solution for detecting outliers.

\noindent\textbf{Constrained Reconstruction as Supervision}
Representation learning for the normal (regular) class under some constraints 
has achieved successes for detecting irregular events in the area of visual data. Therefore, in case that test data does not conform with the imposed constraints, they can potentially be considered as anomaly. As an example, learning the concept of normal class in order to the reconstruct its samples with sparse representation and also minimum effort, which are presented by Cong \etal~\cite{cong2011sparse} and Boiman \etal~\cite{boiman2007detecting} respectively, is widely employed for the task of rare event detection. For more details, Boiman \etal~\cite{boiman2007detecting} consider an event as anomaly if reconstructing it be nearly impossible with accordance to the previous observations. Antic \etal~\cite{antic2011video} proposed a scene parsing approach in which all of the object hypotheses for the foreground of a frame are explained by normal training. Those hypotheses that cannot be explained by normal training are considered as anomaly. In \cite{sabokrou2016video, sabokrou2018adversarially, cong2011sparse}, normal class is learned through a model by reconstructing samples with minimum reconstruction errors. High reconstruction error for a testing sample means the sample is an outlier. Also, \cite{sabokrou2016video,cong2011sparse} introduced a self-representation technique for video anomaly and outlier detection through sparse representation, as a measure for separating inlier and outlier samples. \cite{perera2019ocgan} have exploited GANs  with a constrained latent representations for OCC task. 

In a simple term, we can say that any deviation from an expected and usual behavior is cited as an anomaly in the system.

\section{G2D: Generate to Detect Anomaly}
\label{sec:proposedMethod}
As stated previously, irregularity detection is the task of finding samples with no or low likelihood of occurrence. Abnormal or Irregular samples can be any deviation from an expected and usual behavior or anything which are dissimilar to the target class data \cite{singh2020crowd}. Hence, they are very diverse and researchers prefer to learn the shared concept among the training data, \ie, regular instances. In this case, samples with a different concept are considered as irregularities. In this paper, we follow up a new approach for irregularity detection. Although modeling the outlier samples is difficult, generating them is simple. Due to the high diversity of irregularities, a randomly generated sample can be considered as an outlier instance regarding the target class, with high probability. We Generate to Detect (G2D) anomalies using a binary classification. Our method, \ie, G2D, is composed of three main modules: (1) $\II$rregularity generator network ($\II$), (2) $\CC$ritic network, and (3) $\DD$etector network ($\DD$).  

$\II$ acts as an irregularity generator, while it is totally unaware of such samples. The key idea is to train a GAN on normal samples and utilize its generator before the complete convergence. In fact, generated irregular data should have some deviation from normal instances. Generated irregularities by $\II$ alongside the available normal samples, form an informative training set for optimizing the parameters of a Convolutional Neural Network (CNN) like $\DD$ to distinguish between normal and anomalous samples. A sketch of our proposed method is shown in Fig. \ref{fig:wa}. The detailed information of $\II$, $\CC$ and $\DD$ is described in the following sections.

\subsection{$\II$:~$\II$rregularity Generator Network}
\label{sec:irregularity_generator}
GAN is a well-known framework to implicitly learn the real data distribution. Furthermore, it can effectively generate samples following the learned distribution. Generally, a GAN is composed of two neural networks, Generator (G) and Discriminator (D). G tends to generate samples similar to the real ones, while the D attempts to correctly distinguish between generated (fake) and real (target class) data. These two networks are trained adversarially and in competing with each other through a min-max objective function. There are several efficient methods which take advantage of GANs to learn the distribution of their target class and use D as an irregularity detector \cite{ravanbakhsh2019training,sabokrou2018adversarially,mohammadi2019end,ahmadi2019generative}. Inspired by these works, but from a different point of view, we exploit GANs to learn an efficient deep neural network in presence of samples from both normal and outlier classes. 

As explained previously, owing to the high diversity of samples belonging to the outlier class, researchers are unwilling to model and also learn such classes. Thus, they confine the learning process to only the target class. On the contrary, we aim to generate samples from the unseen class to solve the OCC problem as a simple binary classification task. 

$\II$ is tailored for generating samples from the the unseen class data. This network receives a random noise vector sampled from Gaussian distribution, \ie, $\mathbf{Z} \in \mathcal{N}(\mu_1,\sigma_1)$, as input, and then maps it to an image. Such networks can easily generate samples which do not follow the target class distribution even with no need for the training procedure and only based on randomly initialized parameters. By considering $\mathcal{U}$ and $\mathcal{T}$ as the unseen and target class respectively, the number of unseen samples is very larger than target class data, \ie, $|\mathcal{U}| >> |\mathcal{T}|$ ($|.|$ counts the number of samples). Consequently, $X_r$, as a randomly generated sample by $\II$ belongs to the unseen class with high probability (\ie, $\mathbf{P}(X_r \in \mathcal{U})=\frac{|\mathcal{U}|}{|\mathcal{T}|+|\mathcal{U}|} \approx 1$). Generated instances in this way, is not sufficiently informative to use as training set for a classifier because: (1) generated samples are not semantically similar to the real images, and (2) they are very different from the target class while for a binary classification, samples that are near the boundary of two classes (\eg, support vectors in SVM) are essential. To this end, we train $\II$ on samples from the target class to play the role of G network of GANs. $\II$ and $\CC$ are jointly and adversarially trained. After the training process, $\II$  generates samples following the target class distribution, \ie, $p_{t}$. 

$\II$ network gradually learns the distribution of the target class. Let $p'^i$ be the learned distribution by $\II$ in $i^{th}$ epoch. In the learning period, $p'$ gradually approaches $p_t$. Finally, in the optimum point, \ie, $i=j$, $p'$ meets $p_t$. During the training of $\II$ network, the learned distributions, from $p'^{i=0}$ to $p'^{i=j}$, and generated samples by $\II^{i=1:j}$  can be interpreted as follows: 

\begin{itemize}
	\item $p_t(\II^1(\mathbf{Z})=X_r)=0$ , As explained earlier, we expect that a sample generated by a random CNN, does not comply with the distribution of the target class. Accordingly, samples like $X_r$ are very far from normal instances. 
	
	\item For $k_1<k_2$, $p_t(\II^{i=k_1}(Z))<p_t(\II^{i=k_2}(Z))$. In other words, it is obvious that more training leads to better understanding of real data distribution.
	
	\item We observe that, For $i=k_3$ close to $j$, the KL($p_t(\II^{k_3}(\mathbf{Z})),p_t(\II^j(\mathbf{Z}))<\epsilon$, where KL and $\epsilon$ are the distance between two distributions and a small number close to zero, respectively. In this case, $\II^{k_3}$ generates samples, which in term of texture, are very similar to the target class data, but semantically are completely different. We investigate that such instances can be considered as the boundary outlier class data enclosing the target class samples.  
\end{itemize}

\begin{figure*}
	\centering
	\includegraphics[width=1\linewidth]{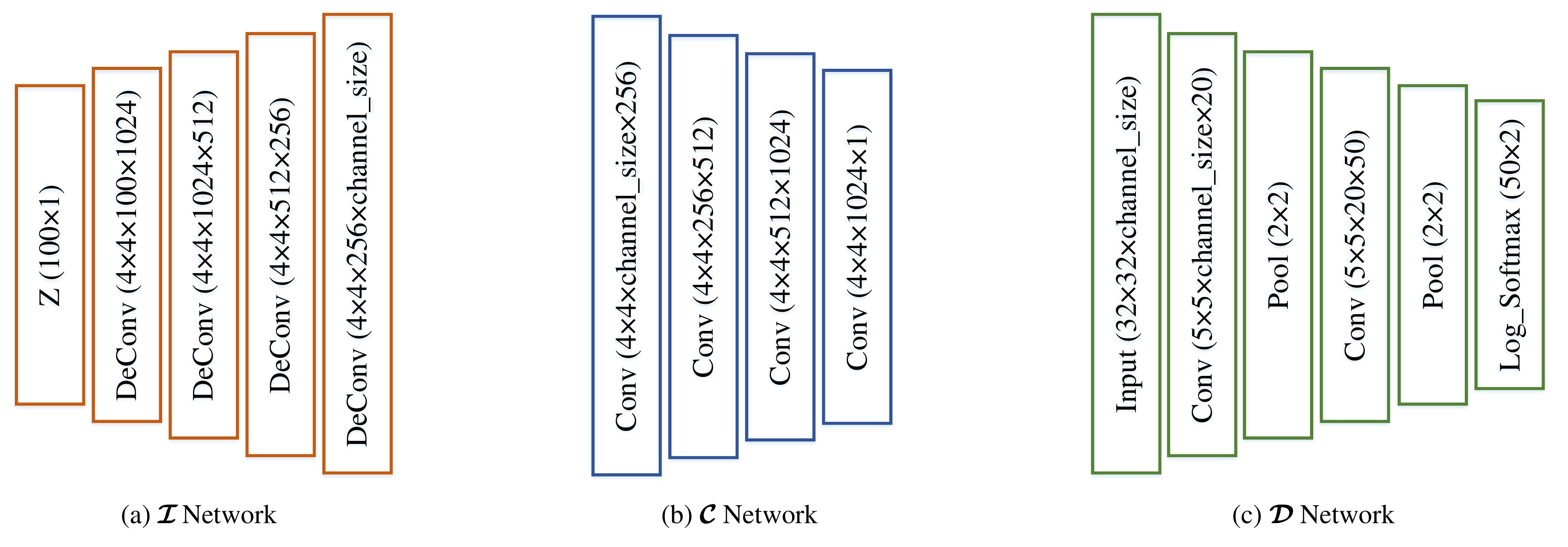}
	\caption{The detailed structure of $\II$, $\CC$ and $\DD$ networks. $\II$ and $\CC$ are adversarially and jointly learned the distribution of the target class data. $\DD$ is trained on normal samples and simulationary generated anomalies. In this figure, the elements of (a, b, c, d, e) show kernel size, kernel size, input channel and output channel, respectively. Note that the channel\_size is equal to 3 for RGB images and 1 for grayscale images}
	\label{fig:networks}
\end{figure*}

In summary, we save the parameters of $\II$ after each epoch. Therefore, we have a set of generator networks, \ie, $\II^{i=1:j}$). $\II^i$ can be used for generating: (1) random samples which are very far from the target class data, (2) outliers surrounding the normal samples, \ie, instances which are in the boundary of the target and the outlier class, and (3) samples following the target class distribution. (1) and (2) are used for generating outlier samples, while (3) is appropriate for augmenting the target class data. By generating outliers, the OCC task turns into a binary classification problem. Note that $\II$ is a CNN which its details is represented in Fig. \ref{fig:networks}.

\subsection{$\DD$:~$\DD$etector Network}
\label{sec:detector}
$\DD$ network is a classifier for distinguishing the normal samples from the simulated anomalies. After the training process (see Section \ref{sec:training}), there are sufficiently available data to train a binary classifier. 
$\mathcal{U} = <\II^{i}(\mathbf{Z}_{1}),~\II^{i}(\mathbf{Z}_{2})~...~\II^{i}(\mathbf{Z}_{m})>,~i \in [1 \cdots k]$ shows the $mk$ samples generated by $\II$ network where $\mathbf{Z} \in \mathcal{N}(\mu_1,\sigma_1)$. Here, $k$ is the number of selected model extracted from the training procedure of the GAN and $m$ represents the number of samples generated by each of selected models. Additionally, 
$\mathcal{T} = <\mathcal{T}_{1}, \mathcal{T}_{2}~...~\mathcal{T}_{n}>$ is the available normal data. Finally, a fully connected neural network, which ends with a softmax layer as a binary classifier, can be trained on $\mathcal{U}$ and $\mathcal{T}$ in order to properly perform the anomaly detection task. The softmax layer outputs two scalar values indicating what class (target or outlier) the processed sample belongs to. The overall scheme of $\DD$ network is represented in Fig. \ref{fig:networks}.  $\DD$ is learned by  optimizing the Equ. \ref{eq:BCLoss}.

\begin{equation}
\label{eq:BCLoss}
\begin{split}
\mathcal{L}_{\text{BC}} = -y\log(p(y))+(1-y)\log(1-p(y))
\end{split}
\end{equation}

Where y is the label (1 for anomaly and 0 for normal) and p(y) is the predicted probability of a sample being abnormal for all data.

\subsection{Training: $\II+\CC$ and $\DD$}
\label{sec:training}
As explained previously, Goodfellow \etal~ \cite{goodfellow2014generative} has introduced an efficient way for adversarial learning of two different neural networks, G and D , called GAN. GANs aim to generate samples that follow the real data distribution through the adversarial training of two networks. G learns to map a latent space like $\mathcal{Z}$ sampled from a specific distribution, i.e., $p_z$, to a real data distribution (referred to as $p_t$). D  is trained by maximizing the probability of assigning the correct label to both the actual data and the fake data from G, while G is simultaneously trained to minimize $\log(1-D(G(Z)))$. In other words, G and D play the following two-players mini-max game:

\begin{equation}
\begin{split}
\underset{G}{\min} ~ \underset{D}{\max} ~~ \Big( & \mathbb{E}_{X \sim p_{t}}[\log(D(X))] \\ 
& + \mathbb{E}_{Z \sim p_{z}}[\log(1-D(G(Z)))]\Big)
\end{split}
\end{equation}

GANs suffer from  the fundamental problems which can also cause difficulties for our algorithm: (1) GANs are not stable and the model may never converge. In fact, the G loss does not indicate that an abnormal sample is far form the normal ones or it resides in the boundary, (2) the behavior of GANs is not interpretative. Generally, we expect that decreasing the loss of G results in increasing the quality of the generated images. But, contrary to our expectation, it may not happen in practice, and (3) mode collapse is common for the conventional GANs.

To avoid that defects of the conventional GANs negatively affect our work, Wasserstein GAN (WGAN) is utilized instead. The discriminator of WGANs are broadly known as $\CC$ritic ($\CC$). $\CC$ network of WGAN is similar to G network of GAN, but the sigmoid function is eliminated. In WGAN, $\CC$ network outputs a scalar score rather than a probability. This score can be interpreted as how real the input images are. In other words, it measures how good a state (the input) is. $\II$ network is optimized based on the Equ. \ref{eq:ILoss}.

\begin{equation}
\label{eq:ILoss}
\mathcal{L}_\II=\frac{1}{m}\sum_{i=1}^{i=m}{f(x^i)-f(\II(Z^i))}
\end{equation}

In GAN, the loss measures how well it fools D networks rather than a measure of the image quality. The G loss in GAN does not drop even the image quality improves. Hence, its value does not properly reflect the progress. Instead, we need to save the testing images and evaluate it visually. On the contrary, the loss function of WGAN indicates the image quality which is more desirable and it is calculated based on the Equ. \ref{eq:CLoss}. 

\begin{equation}
\label{eq:CLoss}
\mathcal{L}_\CC=\frac{1}{m}\sum_{i=1}^{i=m}{f(\II(Z^i))}
\end{equation}

The loss of WGAN is much more stable than GAN loss enabling us to continue learning without experiencing the drop in accuracy. During training, $\II$ models have been saved for each epoch. Accordingly, we have $n$ different models after $n$ epochs. 

Selecting $k$ networks of $n$ saved models as irregularly generators can be done in two ways: (1) using validation set including both normal and anomalous data, and (2) by analyzing $\II$ network loss. In reality, there are not any samples from outlier class. Hence, using validation samples is not feasible, \ie, this approach is flatly contradict our primary assumption. To follow the second solution, the irregularity networks are selected as follows:

\begin{equation}\nonumber
\begin{split}
\begin{array}{c@{\qquad}c}
\label{eq:sparse-formal}
\II(\mathbf{Z})~\text{=}
\begin{cases}
\text{Random irregularity (\ie, noise)}, & \text{if $L<\epsilon_{1} $}.\\
\text{Irregularity close to the boundary}, & \text{if $L<\epsilon_{2} $}.\\
\text{An inlier}, & \text{if $ L<\epsilon_{3} $}.
\end{cases}
\end{array}
\end{split} 
\end{equation}

\begin{figure}
	\centering
	\includegraphics[width=1\columnwidth]{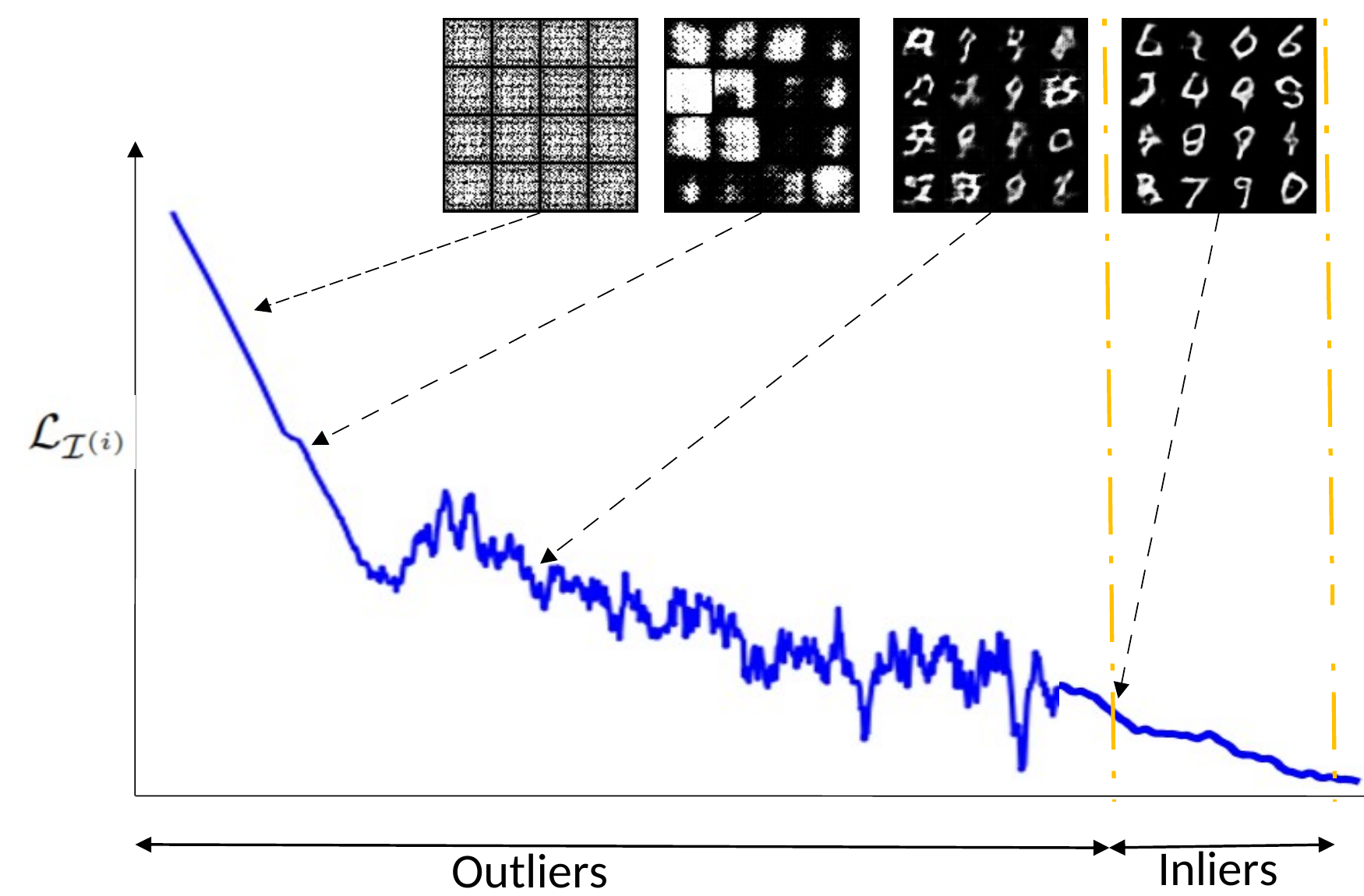}
	\caption{The relation between the loss function values and behavior of $\II$ network.}
	\label{fig:loss_diagram}
\end{figure}

Where,  $L= \mathcal{L}_{\II^{(i)}} - \mathcal{L}_{\II^{(i+h)}}$ (in our case h=5), $\II(\mathbf{Z})$ is a generated irregularity and $\mathcal{L}_{\II^i}$ is the loss value of $\II$ in $i^{th}$ epoch. With respect to Fig. \ref{fig:loss_diagram}, $\epsilon_{1} >> \epsilon_{2}$ and $\epsilon_{2} >> \epsilon_{3}$. Fig \ref{fig:loss_diagram} shows the relation between the loss of $\II$ in training duration and the role of $\II$ network. 

Hereupon, the classifier network should learn the distribution of normal samples to be able to effectively separate normal and abnormal instances from each other. Having selected the $k$ models for $\II$, each of models can generate $m$ samples. In this case, we have $mk$ samples in addition to normal data. For the classifier network, normal samples are considered as real data belonging to the target class while anomalous samples fall into the outlier class because they are considered as the fake data. 

\subsection{Irregularity Detection}
We aim to proposed an end-to-end neural network to detect irregular samples. Accordingly, as explained previously, merely $\DD$ network plays the role of anomaly detector in videos and outlier detector in images. The binary classification problem, which is the converted version of the OCC task in our case, can be simply formulated as:

\begin{equation}\nonumber
\begin{split}
\begin{array}{c@{\qquad}c}
\label{eq:sparse-formal}
\mathcal{D}(X)
\begin{cases}
\text{Normal class (target)}, & \text{if $\DD(X) \geqslant \alpha$}.\\
\text{Abnormal class (irregularity)}, & \text{if $\DD(X) < \alpha$}.
\end{cases}
\end{array}
\end{split} 
\end{equation}

Where $\alpha$ is a predetermined threshold (in our case $\alpha$ is equal to $0.5$).

\section{Experimental Results}
\label{sec:experimentalResults}

We comprehensively evaluate our proposed solution, \ie, G2D, on different datasets. The experimental results alongside an in-depth analysis are presented in this section. Obtained results confirms the superiority of G2D compared to the other baseline and state-of-the-art methods. To show the generality and applicability of the proposed framework for a variety of tasks, we test it for detecting: (1) outlier images, and (2) video anomalies.

\subsection{Implementation Details}
We implement G2D using PyTorch framework and Python
ran on a 64-bit Ubuntu 18.04 LTS system with
an Intel corei7 3.80GHz processor, 16 GB RAM and NVIDIA
GTX 1080 GPU. We use Adam as the optimizer. The learning rate of $\II$ and $\CC$ are the same and equal to 0.0001 with 0.5 as Beta1, 0.999 as Beta2 and a batch size of 64. For $\DD$ network, optimizer is SGD and learning rate is equal to 0.01. Also, momentum is equal to 0.9 with a batch size of 32. We use negative log likelihood function as the loss function for $\DD$ network. The detailed structures of $\II$ and $\DD$ is explained in Sections \ref{sec:irregularity_generator} and \ref{sec:detector}, respectively. These structures are kept fixed for different tasks.

\subsection{Datasets}
\noindent\textbf{UCSD:} This dataset \cite{mahadevan2010anomaly} is composed of two subsets, \textbf{Ped1} and \textbf{Ped2} which are derived from two different outdoor scenes recorded by a static camera with 10 fps. The resolutions of Ped1 and Ped2 are 158$\times$234 and 240$\times$360, respectively. Since the dominant mobile objects are pedestrians in these scenes, the rest objects (\eg, cars, skateboarders, wheelchairs or bicycles) are considered as anomalies. Our proposed algorithm has been evaluated on Ped2.

\noindent\textbf{MNIST:} This dataset \cite{lecun2010mnist} consists of 60,000 handwritten digits from “0” to “9”. Each of digit categories is considered as the target class, \ie, inlier, and the outliers are simulated by randomly sampling images from the other categories with the proportion of 10\% to 50\%. This process has been repeated for all of digit categories.

\noindent\textbf{Caltech-256:} This dataset \cite{griffin2007caltech} includes 256 object categories and 30,607 images in total. Each category contains at least 80 images. Analogous with the previous works \cite{sabokrou2018adversarially}, the procedure is repeated three times. We have used $ n \in {1, 3, 5} $ which are randomly chosen categories as inliers, \ie, target class data. The first 150 images of each category is opted in case that the category has more than 150 images. A certain number of outliers are randomly selected from the “clutter” category. Each of conducted experiments includes exactly 50\% outliers.

\subsection{Video Anomaly Detection}
Visual analysis of unusual and rare events has been drawn the attention of many researchers in computer vision communities. The complexity of the video processing poses even more challenges for detecting anomalous events or novelty compared to outlier detection in image processing. G2D is evaluated on a widely-used and popular dataset, USCD \cite{mahadevan2010anomaly} Ped2. The results are reported based on frame-level basis.

\noindent\textbf{Results on UCSD Ped2:}
In this experiment, each frame is divided into 2D patches of size 30$\times$30 with the overlap of 5. As explained in Section \ref{sec:training}, $\II$ network generates a set of irregular samples with the same size of normal patches. In this way, $\DD$ network is trained on both generated irregular and available normal patches as a binary classifier. In the state of test, $\DD$ is able to detect whether the extracted patches of a frame is abnormal or not. 

The comparison is made based on a frame-level Equal Error Rate (EER). To this end, detecting a pixel of a frame as anomaly leads to considering the whole frame as an abnormal sample. Table \ref{tab:EER} represents the results obtained from the conducted experiments. As it is clear, we have attained comparable and even better result to the other state-of-the-art methods while our work is general with no further tuning to the video based tasks or any other specific applications. Note that G2D treats video as images, \ie, unlike the other approaches (especially Deep-Cascade \cite{sabokrou2017deep} and Deep-Anomaly \cite{sabokrou2018deep}) benefited from both spatial and temporal complex features, our proposed method operates on a patches extracted from the spatial features of the frames.

\begin{table}[t]
	\caption{Frame-level comparisons on Ped2. The best result is boldface and the second best result is underlined.}
	\begin{center}
		\begin{tabular}{llcll}
			\hline
			Method&EER&&Method &EER    \\
			\hline\hline
			MPCCA~\cite{kim2009observe} & 30\%    & &RE  \cite{sabokrou2016video} & 15\%  \\
			\footnotesize{Sabokrou \etal}\cite{sabokrou2018deep}  & 16\%  & &  \footnotesize{Ravanbakhsh \etal}~\cite{ravanbakhsh2019training} &13\%\\
			MDT ~\cite{mahadevan2010anomaly} & 24\% && \footnotesize{Ravanbakhsh \etal}~\cite{ravanbakhsh2017abnormal}& 14\%  \\
			Li \etal\cite{li2013anomaly}  & 18.5\% && Deep-Anomaly\cite{sabokrou2018deep}  & 13\%  \\
			{Dan Xu \etal} ~\cite{xu2014video} &20\%& &  Deep-cascade \cite{sabokrou2019self} &\textbf{9}\% \\
			AVID\cite{sabokrou2018avid}   & 14\% &&    NRE ~\cite{sabokrou2019self} &  {14\%}   \\
			ALOCC \cite{sabokrou2018adversarially} & 13\%&  & Ours (G2D) & \underline{11\%}  \\
			\hline
		\end{tabular}
	\end{center}
	\label{tab:EER}
\end{table}

\subsection{Image Outlier Detection}
If machine learning based methods fail in dealing with processing the data contaminated by noise and irregularities, they experience a considerable drop in performance. The necessity of providing a proper solution to cope with this problem is inevitable because outliers are very common in realistic vision-based training sets. we evaluate the performance of G2D using MNIST and Caltech datasets.

\begin{figure}[t]
	\begin{center}
		\begin{tikzpicture}
		\begin{axis}[width=7.5cm, height=5.5cm,
		symbolic x coords = {10, 20, 30, 40, 50},
		legend pos = south west,
		xlabel={Percentage of outliers (\%)},
		ylabel={$F_1$-Score},
		y label style={at={(axis description cs:0.05,.5)}},
		]
		\addplot+[smooth,blue] coordinates { (10,0.95)(20,0.948)(30,0.942)(40,0.939)(50,0.937)};
		\addplot+[smooth,blue] coordinates { (10,0.97)(20,0.92)(30,0.92)(40,0.91)(50,0.88)};
		\addplot+[smooth,red] coordinates { (10,0.93)(20,0.90)(30,0.87)(40,0.84)(50,0.82)};
		\addplot+[smooth,cyan] coordinates { (10,0.95)(20,0.91)(30,0.88)(40,0.82)(50,0.73)};
		\addplot+[smooth,green] coordinates { (10,0.92)(20,0.83)(30,0.72)(40,0.65)(50,0.55)};
		
		\legend{
			{\footnotesize Ours (G2D) },
			{\footnotesize ALOCC \cite{sabokrou2018adversarially} }, 
			{\footnotesize ALOCC \cite{sabokrou2018adversarially} },
			{\footnotesize DRAE \cite{xia2015learning}}, 
			{\footnotesize LOF \cite{breunig2000lof}}}
		\end{axis}
		\end{tikzpicture}
	\end{center}
	\caption{Comparisons of $F_1$-scores on MNIST dataset for different percentages of outlier samples got involved in the experiment.}
	\label{fig:MNIST}
\end{figure}
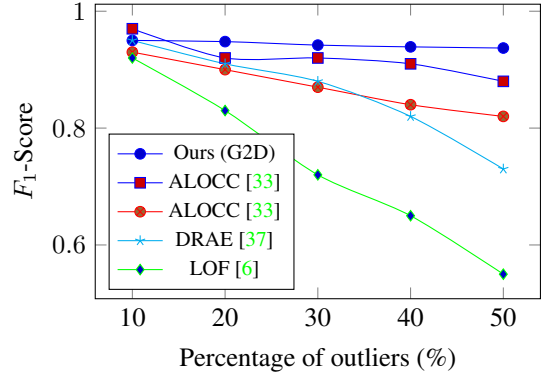

\noindent\textbf{Results on MNIST:}
$\II$ and $\CC$ networks are adversarially and jointly trained on normal samples. Similar to the previous experiments, $mk$ outlier samples simultaneously are generated and $\DD$ network is trained on normal and generated abnormal samples. We report the performance of our method based on $F_1$-score measure against the percentage of outlier instances. Figure \ref{fig:MNIST} confirm that G2D has a better performance, in terms of $F_1$-score, compared to the other state-of-the-art methods. Also, as can be seen, G2D is insensitive to change of outlier portion. The other baseline solutions have a reference model which learns the distribution of normal data. Accordingly, they are able to effectively detect normal sample while facing unusual events cause a sharp drop in their performance. Hence, when outliers form 10\% of the total data, all of the methods performs well. However, with the increase of outliers in number, $F_1$-score has declined. On the other hand, G2D knows the concept of both normal and abnormal data (generated anomalous samples) leading to the higher precision for detecting anomalies and operates more robust. Therefore, the growth of outliers has a negligible effect on the performance of our proposed method and thus it does not experience a noticeable drop in $F_1$-score measure.

\begin{table*}[t]
	\caption{Results on Caltech-256 dataset. Inliers are randomly sampled of one, three or five selected categories. Furthermore, irregularities are randomly chosen from category 257-clutter. \textbf{Two first rows:} Inliers are from one category of images, with $50\%$ portion of outliers; \textbf{Two second rows:} Inliers are from three categories of images, with $50\%$ portion of outliers; \textbf{Two last rows:} Inliers come from five categories of images, while outliers form $50\%$ of the samples.}
	\begin{center}
		\begin{tabular}{ccccccccc}
			\hline 
			& {\footnotesize R-graph \cite{you2017provable}} &   {\footnotesize REAPER \cite{lerman2015robust}} &  {\footnotesize OutlierPursuit \cite{xu2010robust}} &  {\footnotesize LRR \cite{liu2010robust}} &  {\footnotesize SSGAN \cite{kimura2018semi}}  &   {\footnotesize ALOCC \cite{sabokrou2018adversarially} } & {\footnotesize Ours (G2D) }\\
			\hline   \hline  
			AUC  & 0.948 & 0.816 & 0.837 & 0.907 & - & 0.942 & \textbf{0.957}\\
			$F_1$ & 0.914 & 0.808 & 0.823 & 0.893 & 0.977 & 0.928 & \textbf{0.945}\\
			\hline \hline
			AUC  & 0.29 & 0.796  & 0.788 & 0.479 & -   & 0.938 & \textbf{0.951}\\
			$F_1$  & 0.880 & 0.784 & 0.779  & 0.671 & 0.963 &  0.913 & \textbf{0.922}\\
			\hline \hline
			AUC  & 0.913 & 0.657 & 0.629 & 0.337 & -  & 0.923 & \textbf{0.939}\\
			$F_1$  & 0.858  & 0.716 &  0.711  & 0.667 & 0.945  &  0.905 & \textbf{0.913}\\
			\hline
		\end{tabular}
	\end{center}
	\label{tab:caltec}
\end{table*}

\noindent\textbf{Results on Caltech-256:}
We compare our work with other six methods therein designed specifically for outlier detection, including  R-graph \cite{you2017provable}, REAPER \cite{lerman2015robust}, OutlierPursuit \cite{xu2010robust}, LRR \cite{liu2010robust}, SSGAN \cite{kimura2018semi} and ALOCC \cite{sabokrou2018adversarially}. The performance metrics of this experiment are $F_1$-score and the Area Under the Curve (AUC). Results listed in the Table \ref{tab:caltec} confirm G2D performs at least as good as the other solutions, whereas it outperforms them in many cases.

\subsection{Complexity}
In our method, irregularity detection is only performed by $\DD$ network. Hence, there is an end-to-end architecture which is capable of processing an image or video with a lower complexity than the other GAN-based state-of-the-art methods such as AVID\cite{sabokrou2018avid} or ALOCC\cite{sabokrou2018adversarially}.  Irregularity detection is properly done if a sample processed by  two complicated neural networks either for ALOCC or AVD. Otherwise, these algorithms are not able to attain high accuracy. For more details, ALOCC utilized both G network, which includes an encoder-decoder neural network, and D network of a pre-trained GAN to achieve better results.

\section{Discussion}
In this section, the significant issues and the way of dealing with them by our proposed method, \ie, G2D, are thoroughly investigated.

\noindent\textbf{Weak supervision:} G2D is an unsupervised method even though knowing that there is not any irregular samples among the training data can be considered as a weak supervision. In reality, abnormal data are rare and broadly form just a little portion of all samples. Accordingly, the existence of a few number of anomalous instances among the normal data is not problematic in case of considering all the available data as normal.

\noindent\textbf{Generality:} G2D is a general purpose framework  for the OCC task which can easily  fit to various range of problems (see Section \ref{sec:proposedMethod}). However, we have achieved comparable and even better results than  the other state-of-the-arts (see Section \ref{sec:experimentalResults}). Obviously, If we customize our algorithm, \eg, by some modifications in the size and the order of convolutional layers, for a specific-purpose application, the final performance improves.

\noindent\textbf{Convenience of implementation:} G2D is easily implementable forming one of the most important of our contributions. Generally, GAN-based approaches suffer form finding the optimum point for stopping the training process of its two joint neural networks, G and D. In fact, determining a suitable moment in which the discriminator learns the distribution of the normal data is very challenging. On the contrary, our method does not get involved with this burdensome task because we utilized non-optimal trained models for simulationary generating abnormal samples.

\noindent\textbf{Normal data augmentation:} As discussed in Section \ref{sec:irregularity_generator}, all of the models are saved during training. Samples can be generated by the trained models fall into three categories: (1) far from the target class, (2) near to the boundary, and (3) semantically similar to normal samples. Having trained $\II$ network, models belonging to the third category are appropriate for generating samples following the concept of normal data and very similar to them. Therefore, we can use our algorithm for augmenting normal class. This feature is very useful especially when the normal instances are not sufficient.

\noindent\textbf{False positive rate reduction:} The majority of the proposed solution for the OCC task, like anomaly detection, suffer from high false positive rate. This difficulty is mainly caused by the samples resides in the boundary between normal and anomalous data. Lim \etal \cite{lim2018doping} attempted to solve this problem by generating normal data in the boundary. Also, we are able to solve this problem by carefully generating abnormal instances. However, producing anomalies mistakenly does not help. In order to gain better performance in this regard, we can check the quality of simulated anomalies. High quality of the generated samples is an indicator that the model is very close to a suitable boundary which can effectively distinguish between normal and abnormal instances. Note that reducing the false positive rate is just one of the application of our work while it is the main contribution of \cite{lim2018doping}.

\section{Conclusion}
In this paper we propose a general purpose framework for irregularity detection. Firstly, two networks are unsupervisedly and adversarially trained on just normal samples aiming to generate irregular data. Afterward, a deep neural network as a binary classifier learns the simultaneously generated anomalies along with the available normal instances. Result have confirmed that our proposed method, \ie, G2D, can accurately detect irregularities while it is totally unaware of abnormal class data.

\section*{Acknowledgement}
This research was in part supported by a grant from IPM
(No. CS1396-5-01).
{\small
\bibliographystyle{ieee_fullname}
\bibliography{References}

\begin{thebibliography}{10}\itemsep=-1pt

\bibitem{adeli2015robust}
Ehsan Adeli, Kim-Han Thung, Le An, Feng Shi, and Dinggang Shen.
\newblock Robust feature-sample linear discriminant analysis for brain
  disorders diagnosis.
\newblock In {\em Advances in Neural Information Processing Systems}, pages
  658--666, 2015.

\bibitem{ahmadi2019generative}
Milad Ahmadi, Mohammad Sabokrou, Mahmood Fathy, Reza Berangi, and Ehsan Adeli.
\newblock Generative adversarial irregularity detection in mammography images.
\newblock In {\em International Workshop on PRedictive Intelligence In
  MEdicine}, pages 94--104. Springer, 2019.

\bibitem{antic2011video}
Borislav Anti{\'c} and Bj{\"o}rn Ommer.
\newblock Video parsing for abnormality detection.
\newblock In {\em Computer Vision (ICCV), 2011 IEEE International Conference
  on}, pages 2415--2422. IEEE, 2011.

\bibitem{bertini2012multi}
Marco Bertini, Alberto Del~Bimbo, and Lorenzo Seidenari.
\newblock Multi-scale and real-time non-parametric approach for anomaly
  detection and localization.
\newblock {\em Computer Vision and Image Understanding}, 116(3):320--329, 2012.

\bibitem{boiman2007detecting}
Oren Boiman and Michal Irani.
\newblock Detecting irregularities in images and in video.
\newblock {\em International journal of computer vision}, 74(1):17--31, 2007.

\bibitem{breunig2000lof}
Markus~M Breunig, Hans-Peter Kriegel, Raymond~T Ng, and J{\"o}rg Sander.
\newblock Lof: identifying density-based local outliers.
\newblock In {\em ACM sigmod record}, volume~29, pages 93--104. ACM, 2000.

\bibitem{cong2011sparse}
Yang Cong, Junsong Yuan, and Ji Liu.
\newblock Sparse reconstruction cost for abnormal event detection.
\newblock In {\em Computer Vision and Pattern Recognition (CVPR), 2011 IEEE
  Conference on}, pages 3449--3456. IEEE, 2011.

\bibitem{eskin2000anomaly}
Eleazar Eskin.
\newblock Anomaly detection over noisy data using learned probability
  distributions.
\newblock In {\em In Proceedings of the International Conference on Machine
  Learning}. Citeseer, 2000.

\bibitem{goodfellow2014generative}
Ian Goodfellow, Jean Pouget-Abadie, Mehdi Mirza, Bing Xu, David Warde-Farley,
  Sherjil Ozair, Aaron Courville, and Yoshua Bengio.
\newblock Generative adversarial nets.
\newblock In {\em Advances in neural information processing systems}, pages
  2672--2680, 2014.

\bibitem{griffin2007caltech}
Gregory Griffin, Alex Holub, and Pietro Perona.
\newblock Caltech-256 object category dataset.
\newblock 2007.

\bibitem{kim2009observe}
Jaechul Kim and Kristen Grauman.
\newblock Observe locally, infer globally: a space-time mrf for detecting
  abnormal activities with incremental updates.
\newblock In {\em 2009 IEEE Conference on Computer Vision and Pattern
  Recognition (CVPR)}, pages 2921--2928. IEEE, 2009.

\bibitem{kimura2018semi}
Masanari Kimura and Takashi Yanagihara.
\newblock Semi-supervised anomaly detection using gans for visual inspection in
  noisy training data.
\newblock {\em arXiv preprint arXiv:1807.01136}, 2018.

\bibitem{lecun2010mnist}
Yann LeCun, Corinna Cortes, and Christopher~JC Burges.
\newblock Mnist handwritten digit database.
\newblock {\em AT\&T Labs [Online]. Available: http://yann. lecun.
  com/exdb/mnist}, 2, 2010.

\bibitem{lerman2015robust}
Gilad Lerman, Michael~B McCoy, Joel~A Tropp, and Teng Zhang.
\newblock Robust computation of linear models by convex relaxation.
\newblock {\em Foundations of Computational Mathematics}, 15(2):363--410, 2015.

\bibitem{li2013anomaly}
Weixin Li, Vijay Mahadevan, and Nuno Vasconcelos.
\newblock Anomaly detection and localization in crowded scenes.
\newblock {\em IEEE transactions on pattern analysis and machine intelligence},
  36(1):18--32, 2013.

\bibitem{lim2018doping}
Swee~Kiat Lim, Yi Loo, Ngoc-Trung Tran, Ngai-Man Cheung, Gemma Roig, and Yuval
  Elovici.
\newblock Doping: Generative data augmentation for unsupervised anomaly
  detection with gan.
\newblock In {\em 2018 IEEE International Conference on Data Mining (ICDM)},
  pages 1122--1127. IEEE, 2018.

\bibitem{liu2010robust}
Guangcan Liu, Zhouchen Lin, and Yong Yu.
\newblock Robust subspace segmentation by low-rank representation.
\newblock In {\em Proceedings of the 27th international conference on machine
  learning (ICML-10)}, pages 663--670, 2010.

\bibitem{mahadevan2010anomaly}
Vijay Mahadevan, Weixin Li, Viral Bhalodia, and Nuno Vasconcelos.
\newblock Anomaly detection in crowded scenes.
\newblock In {\em 2010 IEEE Computer Society Conference on Computer Vision and
  Pattern Recognition (CVPR)}, pages 1975--1981. IEEE, 2010.

\bibitem{markou2003novelty}
Markos Markou and Sameer Singh.
\newblock Novelty detection: a review—part 1: statistical approaches.
\newblock {\em Signal processing}, 83(12):2481--2497, 2003.

\bibitem{mohammadi2019end}
Bahram {Mohammadi} and Mohammad {Sabokrou}.
\newblock End-to-end adversarial learning for intrusion detection in computer
  networks.
\newblock In {\em 2019 IEEE 44th Conference on Local Computer Networks (LCN)},
  pages 270--273, 2019.

\bibitem{morris2011trajectory}
Brendan~Tran Morris and Mohan~M Trivedi.
\newblock Trajectory learning for activity understanding: Unsupervised,
  multilevel, and long-term adaptive approach.
\newblock {\em IEEE transactions on pattern analysis and machine intelligence},
  33(11):2287--2301, 2011.

\bibitem{nguyen2019anomaly}
Trong-Nguyen Nguyen and Jean Meunier.
\newblock Anomaly detection in video sequence with appearance-motion
  correspondence.
\newblock In {\em Proceedings of the IEEE International Conference on Computer
  Vision}, pages 1273--1283, 2019.

\bibitem{perera2019ocgan}
Pramuditha Perera, Ramesh Nallapati, and Bing Xiang.
\newblock Ocgan: One-class novelty detection using gans with constrained latent
  representations.
\newblock In {\em Proceedings of the IEEE Conference on Computer Vision and
  Pattern Recognition}, pages 2898--2906, 2019.

\bibitem{rahmani2016coherence}
Mostafa Rahmani and George Atia.
\newblock Coherence pursuit: Fast, simple, and robust principal component
  analysis.
\newblock {\em arXiv preprint arXiv:1609.04789}, 2016.

\bibitem{ravanbakhsh2017abnormal}
Mahdyar Ravanbakhsh, Moin Nabi, Enver Sangineto, Lucio Marcenaro, Carlo
  Regazzoni, and Nicu Sebe.
\newblock Abnormal event detection in videos using generative adversarial nets.
\newblock In {\em 2017 IEEE International Conference on Image Processing
  (ICIP)}, pages 1577--1581. IEEE, 2017.

\bibitem{ravanbakhsh2019training}
Mahdyar Ravanbakhsh, Enver Sangineto, Moin Nabi, and Nicu Sebe.
\newblock Training adversarial discriminators for cross-channel abnormal event
  detection in crowds.
\newblock In {\em 2019 IEEE Winter Conference on Applications of Computer
  Vision (WACV)}, pages 1896--1904. IEEE, 2019.

\bibitem{sabokrou2016video}
Mohammad Sabokrou, Mahmood Fathy, and Mojtaba Hoseini.
\newblock Video anomaly detection and localisation based on the sparsity and
  reconstruction error of auto-encoder.
\newblock {\em Electronics Letters}, 52(13):1122--1124, 2016.

\bibitem{sabokrou2015real}
Mohammad Sabokrou, Mahmood Fathy, Mojtaba Hoseini, and Reinhard Klette.
\newblock Real-time anomaly detection and localization in crowded scenes.
\newblock In {\em Proceedings of the IEEE CVPR Workshops}, pages 56--62, 2015.

\bibitem{sabokrou2020deep}
Mohammad Sabokrou, Mahmood Fathy, Guoying Zhao, and Ehsan Adeli.
\newblock Deep end-to-end one-class classifier.
\newblock {\em IEEE Transactions on Neural Networks and Learning Systems},
  2020.

\bibitem{sabokrou2017deep}
Mohammad Sabokrou, Mohsen Fayyaz, Mahmood Fathy, and Reinhard Klette.
\newblock Deep-cascade: Cascading 3d deep neural networks for fast anomaly
  detection and localization in crowded scenes.
\newblock {\em IEEE Transactions on Image Processing}, 26(4):1992--2004, 2017.

\bibitem{sabokrou2018deep}
Mohammad Sabokrou, Mohsen Fayyaz, Mahmood Fathy, Zahra Moayed, and Reinhard
  Klette.
\newblock Deep-anomaly: Fully convolutional neural network for fast anomaly
  detection in crowded scenes.
\newblock {\em Computer Vision and Image Understanding}, 172:88--97, 2018.

\bibitem{sabokrou2019self}
Mohammad Sabokrou, Mohammad Khalooei, and Ehsan Adeli.
\newblock Self-supervised representation learning via neighborhood-relational
  encoding.
\newblock {\em International Conference on Computer Vision (ICCV)}, 2019.

\bibitem{sabokrou2018adversarially}
Mohammad Sabokrou, Mohammad Khalooei, Mahmood Fathy, and Ehsan Adeli.
\newblock Adversarially learned one-class classifier for novelty detection.
\newblock In {\em Proceedings of the IEEE Conference on Computer Vision and
  Pattern Recognition (CVPR)}, pages 3379--3388, 2018.

\bibitem{sabokrou2018avid}
Mohammad Sabokrou, Masoud Pourreza, Mohsen Fayyaz, Rahim Entezari, Mahmood
  Fathy, J{\"u}rgen Gall, and Ehsan Adeli.
\newblock Avid: adversarial visual irregularity detection.
\newblock In {\em Asian Conference on Computer Vision}, pages 488--505.
  Springer, 2018.

\bibitem{singh2020crowd}
Kuldeep Singh, Shantanu Rajora, Dinesh~Kumar Vishwakarma, Gaurav Tripathi,
  Sandeep Kumar, and Gurjit~Singh Walia.
\newblock Crowd anomaly detection using aggregation of ensembles of fine-tuned
  convnets.
\newblock {\em Neurocomputing}, 371:188 -- 198, 2020.

\bibitem{tang2020integrating}
Yao Tang, Lin Zhao, Shanshan Zhang, Chen Gong, Guangyu Li, and Jian Yang.
\newblock Integrating prediction and reconstruction for anomaly detection.
\newblock {\em Pattern Recognition Letters}, 129:123 -- 130, 2020.

\bibitem{xia2015learning}
Yan Xia, Xudong Cao, Fang Wen, Gang Hua, and Jian Sun.
\newblock Learning discriminative reconstructions for unsupervised outlier
  removal.
\newblock In {\em Proceedings of the IEEE International Conference on Computer
  Vision}, pages 1511--1519, 2015.

\bibitem{xu2015learning}
Dan Xu, Elisa Ricci, Yan Yan, Jingkuan Song, and Nicu Sebe.
\newblock Learning deep representations of appearance and motion for anomalous
  event detection.
\newblock {\em arXiv preprint arXiv:1510.01553}, 2015.

\bibitem{xu2014video}
Dan Xu, Rui Song, Xinyu Wu, Nannan Li, Wei Feng, and Huihuan Qian.
\newblock Video anomaly detection based on a hierarchical activity discovery
  within spatio-temporal contexts.
\newblock {\em Neurocomputing}, 143:144--152, 2014.

\bibitem{xu2010robust}
Huan Xu, Constantine Caramanis, and Sujay Sanghavi.
\newblock Robust pca via outlier pursuit.
\newblock In {\em Advances in Neural Information Processing Systems}, pages
  2496--2504, 2010.

\bibitem{yamanishi2004line}
Kenji Yamanishi, Jun-Ichi Takeuchi, Graham Williams, and Peter Milne.
\newblock On-line unsupervised outlier detection using finite mixtures with
  discounting learning algorithms.
\newblock {\em Data Mining and Knowledge Discovery}, 8(3):275--300, 2004.

\bibitem{you2017provable}
Chong You, Daniel~P Robinson, and Ren{\'e} Vidal.
\newblock Provable self-representation based outlier detection in a union of
  subspaces.
\newblock In {\em Proceedings of the IEEE Conference on Computer Vision and
  Pattern Recognition}, pages 3395--3404, 2017.

\end{thebibliography}
}

\end{document}